\documentclass{article}
\usepackage{ijcai22}

\usepackage{times}
\usepackage{soul}
\usepackage{url}
\usepackage[hidelinks]{hyperref}
\usepackage[utf8]{inputenc}
\usepackage[small]{caption}
\usepackage{graphicx}
\usepackage{amsmath}
\usepackage{amsthm}
\usepackage{booktabs}
\usepackage{algorithm}
\usepackage{algorithmic}
\usepackage{amsfonts}
\usepackage{multirow}
\usepackage{subfigure}

\title{Label Noise-Resistant Mean Teaching for Weakly Supervised Fake News Detection}

\author{
Jingyi Xie\textsuperscript{\rm \dag}, 
Jiawei Liu\textsuperscript{\rm \dag}, 
Zheng-Jun Zha\textsuperscript{\rm *}
\affiliations{
    University of Science and Technology of China, China \\
    hsfzxjy@mail.ustc.edu.cn, \{jwliu6,zhazj\}@ustc.edu.cn
}}

\begin{document}

\maketitle


\begin{abstract}
   Fake news spreads at an unprecedented speed, reaches global audiences and poses huge risks to users and communities. Most existing fake news detection algorithms focus on building supervised training models on a large amount of manually labeled data, which is expensive to acquire or often unavailable. In this work, we propose a novel label noise-resistant mean teaching approach (LNMT) for weakly supervised fake news detection. LNMT leverages unlabeled news and feedback comments of users to enlarge the amount of training data and facilitates model training by generating refined labels as weak supervision. Specifically, LNMT automatically assigns initial weak labels to unlabeled samples based on semantic correlation and emotional association between news content and the comments. Moreover, in order to suppress the noises in weak labels, LNMT establishes a mean teacher framework equipped with label propagation and label reliability estimation. The framework measures a weak label similarity matrix between the teacher and student networks, and propagates different valuable weak label information to refine the weak labels. Meanwhile, it exploits the consistency between the output class likelihood vectors of the two networks to evaluate the reliability of the weak labels and incorporates the reliability into model optimization to alleviate the negative effect of noisy weak labels. Extensive experiments show the superior performance of LNMT.
\end{abstract}

\section{Introduction}

\footnotetext{\dag \ Equal contribution.}
\footnotetext{* Corresponding author.}

In recent years, the boom of social media has created a straight path from content producers to content consumers, changing the way that people acquire information, express feelings and exchange opinions \cite{PreferenceAware}. However, such a convenient platform also promotes the wide dissemination of fake news (false or misleading information dissimulated in news articles to mislead people) as it enables people to create and spread any information at any time with a low cost. The dissemination of fake news may cause many negative effects, including social panics and financial losses. For example, fake news about the COVID-19 pandemic can prevent healthy behaviors and increase the spread of the virus \cite{lyu2021covid}. Thus, automatic fake news detection has become an urgent problem attracting much research effort.



Automatic fake news detection aims to design a classifier model to identify a given news as real or fake \cite{tschiatschek2018fake}. Early studies mainly focus on traditional machine learning models based on feature engineering \cite{jin2017detection}. After the emergence of deep learning technique, many deep learning based methods \cite{bian2020rumor,khoo2020interpretable,jin2017MM} have been proposed and greatly improved the performance, due to the powerful ability of automatically learning informative representations from samples. Nevertheless, these deep learning based methods rely on large amount of labeled samples to train supervised models. Creating such large training data is extremely expensive and time-consuming, requiring the annotators to have sufficient knowledge about the events. To fully utilize the ability of deep learning models in fake news detection, it is essential to tackle the problem of labeling fake news.

A promising solution is to exploit the feedback reports of users (\textit{i.e.}, comments) who read the news article. Most popular social medial platforms offer a way for users to report their comments about the news article, which contain significant information that are highly relevant to fake news detection \cite{shu2019beyond,qian2018neural}, in the form of opinions, stances, and emotion, \textit{etc}. The reports of users can be served as weak sources of labels, which bear implicit judgments of the users about the news, and are more easily available. The large amount of users' reports help alleviate the annotation shortage problem for fake news detection. Shu \textit{et al.} \cite{shu2020leveraging} proposes to exploit multiple weak supervision signals from different sources from user engagements with the news content to generate weak labels for unlabeled samples, and jointly leverage the limited labeled samples along with these weakly labeled samples to train a fake news detector in a meta-learning framework. However, the generated weak labels are unavoidably noisy due to inaccurate prediction model and the meaningless or poor-quality comments, which hinder model learning and performance improvement. Therefore, how to obtain refined weakly labeled samples is a major issue. Wang \textit{et al.} \cite{wang2020weak} attempts to address this issue, and proposes a preliminary reinforced weakly-supervised framework (WeFEND) that uses reinforcement learning technique to directly select high-quality samples from the estimated weak labels to train the fake news detector. However, in WeFEND, the annotator assigns weak labels simply based on user reports and is trained in an off-line manner. It is independent from the training process of the reinforced selector and the fake news detector, therefore the reliability of the estimated weak labels cannot be improved by themselves and a large number of inaccurate weakly labeled samples are directly discarded from the subsequent training process of fake news detector. Moreover, WeFEND neglects the emotion association between news content and the feedback reports, which contain complementary and useful clues for fake news detection \cite{giachanou2019leveraging,zhang2021mining}.

\begin{figure*}[t]
    \centering
    \includegraphics[page=1,width=1.0\linewidth]{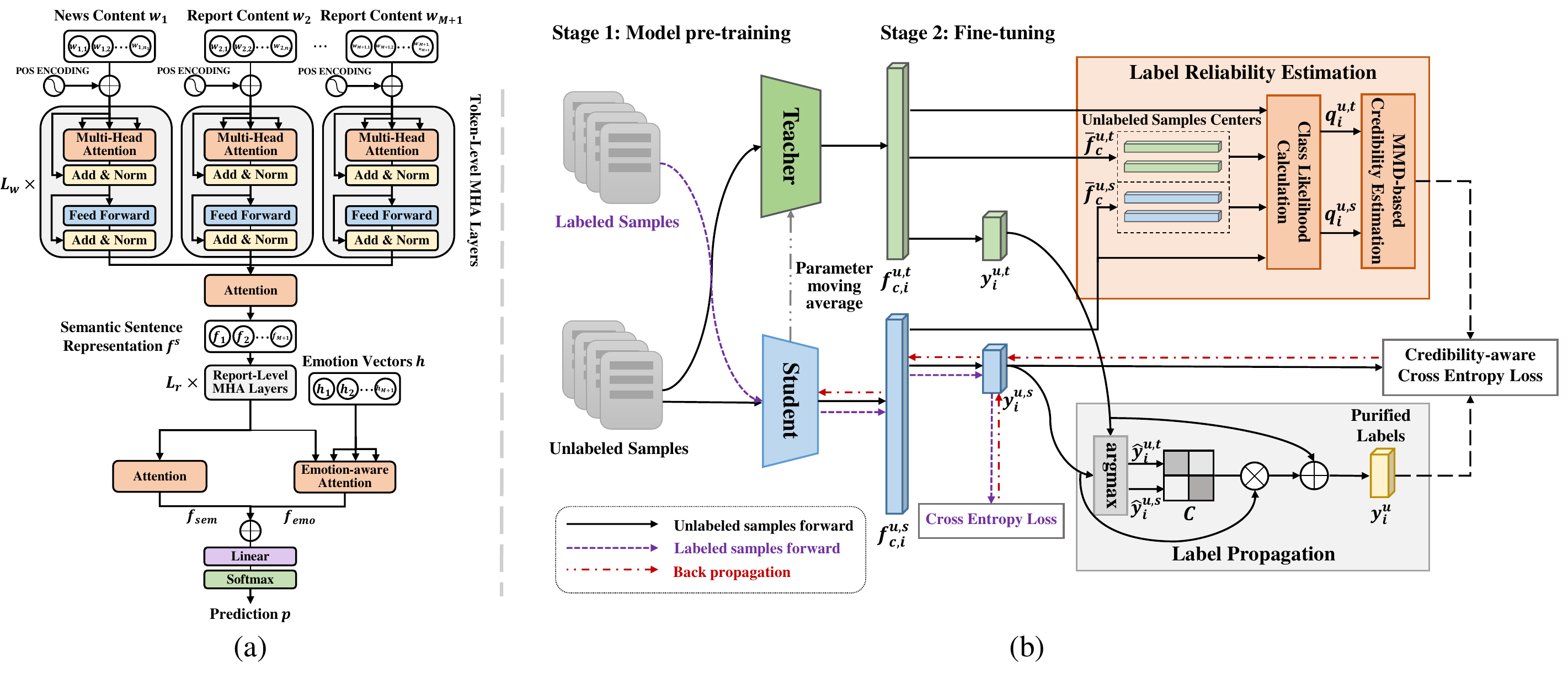}
    \caption{The overall architecture of the proposed LNMT. (a) The detailed structure of the baseline model. (b) The detailed structure of the mean teacher framework equipped with label propagation and label reliability estimation strategies.}
    \label{fig:pipeline}
\end{figure*}

In this work, we propose a novel label noise-resistant mean teaching approach (LNMT) for weakly supervised fake news detection,  which leverages unlabeled news and feedback comments of users to enlarge the amount of training data and facilitates model training by generating refined labels as weak supervision. Specifically, as illustrated in Figure~\ref{fig:pipeline}, LNMT establishes a mean teacher framework with different learning ability and information interaction. The framework consists of a teacher network and a student network that are based on the same baseline model. The training process of the mean teacher framework has two stages. In the first training stage, we pre-train a baseline model on the limited manually labeled samples, which is then used as the initialization of the teacher and student networks, and automatically assigns initial weak labels for unlabeled samples. The baseline model simultaneously explores the semantic correlation and the emotional association between news content and the feedback reports for enhancing the quality of the initial weak labels. In the second training stage, in order to depress noises in the weak labels, the mean teacher framework measures weak label similarity matrix of the unlabeled samples between the two networks, and propagates different valuable weak label information encoded in the two networks to refine the weak labels. Meanwhile, the framework exploits the consistency between the output class likelihood vectors of the two networks to evaluate the reliability of weak labels, and incorporates the reliability into model optimizing for alleviating the negative effect on noisy weak labels. By jointly performing the label propagation (LP) and label reliability estimation (LR) strategies, the framework can effectively leverage the limited manually labeled samples and vast refined weakly labeled samples to train a satisfactory fake news detector. Extensive experiments on a benchmark dataset show the effectiveness of the proposed approach.

The main contributions of this paper are as follows: 
(1) Considering the shortage of high-quality labels, we propose a novel label noise-resistant mean teaching  (LNMT) method for weakly supervised fake news detection. (2) LNMT simultaneously explores semantic correlation and emotional association in news and user reports to assign initial weak labels with reasonable quality. (3) With the proposed label propagation and label reliability estimation strategies, LNMT has the ability to refine weak labels and alleviate the negative effect of noisy weak labels.



\section{Related Work}

\textbf{Supervised Fake News Detection.} Most existing methods mainly focus on extracting rumor features from the text contents, user profiles and propagation structures, and training supervised classifiers on labeled samples \cite{kwon2013prominent,zhao2015enquiring}. Early traditional works mainly design a complementary set of hand-crafted features. For example,  Kwon \textit{et al.} \cite{kwon2013prominent} studied the rumor spreading pattern on Twitter by exploring three sets of features: temporal, structural and linguistic. Recently, researchers exploit deep learning models that automatically mine high-level representation to identify fake news \cite{khoo2020interpretable,yu2017convolutional}. For example, Wang \textit{et al.} \cite{Wang2021M} proposed a detection framework named MetaFEND, which can learn quickly with a few verified posts using the proposed meta neural process networks for effective detection on newly arrived events. 

\noindent\textbf{Weakly Supervised Fake News Detection.} Although supervised methods have made a remarkable breakthrough on performance, they rely on large amount of labeled samples, which are extremely expensive to obtain. To handle this challenge, learning with weak supervision provides a promising solution. For eaxmple, Shu \textit{et al.} \cite{shu2020leveraging} leveraged a small amount of labeled data and multi-source weak signals from social engagements with their importance weights, to train a fake news detector in a meta-learning framework. Wang \textit{et al.} \cite{wang2020weak} proposed a reinforced weakly-supervised fake news detection framework, which consists of an annotator, a reinforced selector and a fake news detector. The annotator assigned weak labels for unlabeled samples based on user reports. The reinforced selector employed reinforcement learning technique to select high-quality samples from weakly labeled samples. The fake news detector identified fake news based on the news content. 

\section{Method}




\subsection{Architecture Overview}
Fake news detection task can be defined as a binary classification problem, \textit{i.e.}, each news article can be real ($\boldsymbol{y} = 0$) or fake ($\boldsymbol{y} = 1$). Let $\{\boldsymbol{X}^l, \boldsymbol{Y}^l\} = \{(\boldsymbol{x}_i^l,\boldsymbol{y}_i^l)\}_{i=1}^{N_l}$ denotes $N_l$ manually annotated news article $\boldsymbol{x}_i^l$ with the associated label $\boldsymbol{y}_i^l$, and $\boldsymbol{X}^u = \{\boldsymbol{x}_i^u\}_{i=1}^{N_u}$ denotes $N_u$ unlabeled news article $\boldsymbol{x}_i^u$. In addition, each news article $\boldsymbol{x}_i$ contains $M_i$ feedback reports of users $\boldsymbol{R}_i=\{\boldsymbol{r}_{i,1},..., \boldsymbol{r}_{i,M_i}\}$. Given the limited amount of labeled samples and large amount of unlabeled samples, the objective is to effectively utilize the unlabeled data as weak supervision to improve the performance. To this end, we propose a novel label noise-resistant mean teaching approach (LNMT) for weakly supervised fake news detection. The overall architecture is shown in Figure~\ref{fig:pipeline}. It establishes a mean teacher framework equipped with label propagation and label reliability estimation strategies. The framework consists of a teacher network and a student network, both of which are based on a baseline model. The training process of the framework have two stages, which jointly leverages the limited labeled samples and vast refined weakly labeled samples to train an effective fake news detector.



\subsection{The Baseline Model}
The baseline model is the core component of the overall framework of LNMT, which aims at automatically assigning initial weak labels to the unlabeled samples, and enlarging the training set. It contains a transformer encoder and an emotion-aware encoder. We employ the feedback reports provided by users as weak supervision signals to train the baseline model in the first training stage.

\textbf{Transformer Encoder.} Transformer \cite{vaswani2017attention} has a strong ability to extract semantic features by modeling long-term dependency and capturing the sequence information through positional encoding. We thus design a transformer encoder as the core of the baseline model to explore the semantic interaction between news content and the reports, and learn the semantic feature $\boldsymbol{f}_{sem}$. The building block of the transformer encoder is the multi-head attention (MHA) layer \cite{vaswani2017attention}. An MHA layer consists of a multi-head self-attention module and a fully connected feed-forward network. 


Given a sample consisting of a news article $\boldsymbol{x}$ and $M$ reports $\boldsymbol{R}=\{\boldsymbol{r}_{i}\}_{i=1}^{M}$, the sample can be denoted as $\mathcal{W}= [\boldsymbol{w}_1, \boldsymbol{w}_2,..., \boldsymbol{w}_{M+1}]$, where $\boldsymbol{w}_i$ denotes $i$-th sequence text (corresponding to the news article or each report). Each sequence text $\boldsymbol{w}_i = [\boldsymbol{w}_{i,1}, \boldsymbol{w}_{i,2},..., \boldsymbol{w}_{i,n_i}]$ have $n_i$ words. The words in the news article and the reports are firstly projected into continuous word embeddings. The sequences of word embeddings in the news article and the reports are passed through $L_w$ layers of token-level MHA to produce the complex sequence representations $\{\boldsymbol{z}_{i,j}\}_{i=1}^{M+1}$. These layers allow information interaction between tokens in a news article or a report. The sequence features $\{\boldsymbol{z}_{i,j}\}_{i=1}^{M+1}$ are then aggregated by attention mechanism to generate the semantic sentence representations $\{\boldsymbol{f}_i^s\}_{i=1}^{M+1}$ for the news article and the reports. $\boldsymbol{f}_i^s$ is calculated as following:
\begin{align}
   \{\boldsymbol{z}_{i,j}\}_{j=1}^{n_i} & =\text{MHA}_{L_w}(\{\boldsymbol{w}_{i,j}\}_{j=1}^{n_i})      \notag  \\
    g_{i,j}      & =\text{Softmax}(\boldsymbol{\gamma}_w^T\boldsymbol{z}_{i,j})  \notag \\
    \boldsymbol{f}_i^s  & =\sum_{j=1}^{n_i}g_{ij}\boldsymbol{z}_{i,j}
    \label{eq1}
\end{align}
where $\boldsymbol{w}_{i,j}$ denotes $j$-th word tokens of $i$-th sequence text, $n_i$ is the number of word tokens for $i$-th sequence text. $\boldsymbol{\gamma}_w \in  \mathbb{R}^{d_{model}}$ denotes the learnable training parameter of the linear transformation layer and $g_{ij}$ denotes the importance scores of the word tokens. Furthermore, the obtained sentence representations $\{\boldsymbol{f}_i^s\}_{i=1}^{M+1}$ are fed into extra $L_r$ layers of report-level MHA, which model the information interactions between the news article and the reports at report-level. Lastly, the attention mechanism is used to interpolate the news article and the reports to obtain the final semantic representation $\boldsymbol{f}_{sem}$ of the sample, which is calculated as following:
\begin{align}
    \{\boldsymbol{z}_{i}^s\}_{i=1}^{M+1}  & =\text{MHA}_{L_r}(\{\boldsymbol{f}_{i}^s\}_{i=1}^{M+1})   \notag    \\
    \pi_{i}  & =\text{Softmax}(\boldsymbol{\gamma}_s^T \boldsymbol{z}_{i}^s)                            \notag \\
   \boldsymbol{f}_{sem} & =\sum_{i=1}^{M+1}\pi_{i} \boldsymbol{z}_{i}^s                        
\end{align}
where $\boldsymbol{f}_i^s\in \mathbb{R}^d$, and $\boldsymbol{\gamma}_s \in \mathbb{R}^{d_{model}}$ denotes the learnable training parameter of the linear transformation layer. 



\textbf{Emotion-aware Encoder.} Prior works \cite{giachanou2019leveraging,ajao2019sentiment} have shown that there exists obvious relationship between news veracity and the emotion information in the reports of users. Following this idea, we design an emotion-aware encoder, which mines the emotion information contained in the news article and the user reports, and explores the emotional association between them to further enhance feature representation. Specifically, the emotion-aware encoder firstly extracts emotion vectors $\{\boldsymbol{h}_i\} \in \mathbb{R}^{d_{emo}}$ from the news article and the reports. It then projects $\boldsymbol{h_i}$ into three different representation subspaces $\boldsymbol{h}_i^{(q)}$, $\boldsymbol{h}_i^{(k)}$ and $\boldsymbol{h}_i^{(v)}$, and applies an emotion-aware attention module (at the report-level) with $\boldsymbol{h}^{(q)}_i=\boldsymbol{h}^{(q)}_i+\boldsymbol{z}_i^s$ as queries, $\boldsymbol{h}_i^{(k)}$ as keys, and $\boldsymbol{h}_i^{(v)}$ as values, to generate the final emotion feature $\boldsymbol{f}_{emo}$ for the sample. The feature $\boldsymbol{f}_{emo}$ is calculated as following:
\begin{align}
    \boldsymbol{h}_i^{(q)} & =\boldsymbol{W}_{hq}^T\boldsymbol{h}_i+\boldsymbol{z}_i^s, \;
    \boldsymbol{h}_i^{(k)}  =\boldsymbol{W}_{hk}^T\boldsymbol{h}_i, \;  \notag  \boldsymbol{h}_i^{(v)}=\boldsymbol{W}_{hv}^T\boldsymbol{h}_i \notag             
     \\
    \beta_{ji}  & =\text{Softmax}(\frac{\boldsymbol{h}^{(q)}_i\boldsymbol{h}^{(k)T}_j}{\sqrt{d_{model}}})  \notag \\
    \boldsymbol{f}_{emo} & =\frac{1}{M+1}\sum_{j=1}^{M+1}\sum_{i=1}^{M+1}\beta_{ji}\boldsymbol{h}_i^{(v)}
\end{align}
where $\boldsymbol{W}_{hq},\boldsymbol{W}_{hk}, \boldsymbol{W}_{hv} \in \mathbb{R}^{d_{emo} \times d_{model}}$ are the weight matrices of linear projections. $\beta_{ji}$ denotes emotion-aware attention, which measures the pairwise compatibility of the emotion features. We concatenate the complementary two features $\boldsymbol{f}_{c}=\boldsymbol{f}_{sem} \oplus\boldsymbol{f}_{emo}$ to produce the reinforced semantic feature $\boldsymbol{f}_{c}$. The feature $\boldsymbol{f}_{c}$ is then fed into a classifier layer for prediction. In the first training stage, we employ a binary cross entropy loss to train the baseline model with the labeled samples under supervised setting. 



\subsection{Mean Teacher Framework}
Given the unlabeled set $\boldsymbol{X}^u$ with the corresponding reports, the pre-trained baseline model can predict their weak labels, denoted as $\hat{\boldsymbol{Y}}^u$. Nevertheless, the automatically annotated weak labels are unavoidably noisy and directly utilizing these samples $\{\boldsymbol{X}^u, \hat{\boldsymbol{Y}}^u\}$ to train the model could mislead feature learning and degrade the performance. The mean teacher framework is designed to refine the weak labels by utilizing label propagation, and evaluate the reliability of them for alleviating the negative effect on noisy weakly labeled samples during modeling optimization. 

The mean teacher framework consists of a teacher network $F_t$ and a student network $F_s$, both of which are initialized by the pre-trained baseline model in the first training stage. The teacher network $F_t$ acts as a teacher that periodically produces soft weak labels for unlabeled samples, which are employed as weak supervision signals for the student network $F_s$. In the second training stage, the parameters of $F_s$ are updated through back-propagation, while $F_t$ are updated through exponential moving average policy on the parameters of $F_s$. This policy guarantees the two networks have different learning ability and label evolving is stable \cite{tarvainen2017mean}. The second training stage has $T$ training generations (epochs). Before each generation, we feed all unlabeled samples into $F_t$, and obtain the soft predictions $\{\boldsymbol{y}_i^u = \boldsymbol{p}_i^{u,t}\}_{i=1}^{N_u}$, $\boldsymbol{y}_i^u \in [0,1]$ will serve as the soft weak label for sample $\boldsymbol{x}_i^u$ in the next generation. During each training generation, the loss function $\mathcal{L}_{s}$ for the student network $F_s$ is formulated as following:
\begin{align}
    \mathcal{L}_l &= \frac{1}{B}\sum_{i=1}^B -[\boldsymbol{y}_i^l \log \boldsymbol{p}_i^l
    + (1-\boldsymbol{y}_i^l) \log (1-\boldsymbol{p}_i^l)]                                                  \notag            \\
    \mathcal{L}_u &= \frac{1}{B}\sum_{i=1}^B -[\boldsymbol{y}_i^u \log \boldsymbol{p}_i^{u,s} 
    + (1-\boldsymbol{y}_i^u) \log (1-\boldsymbol{p}_i^{u,s})]                                   \notag \\
    \mathcal{L}_{s} &= \mathcal{L}_l + \mathcal{L}_u
    \label{loss}
\end{align}
where $\mathcal{L}_l,\mathcal{L}_u$ are the losses for labeled and unlabeled samples respectively. $\boldsymbol{p}_i^l$, $\boldsymbol{p}_i^{u,s}$ are the soft prediction scores of the labeled and unlabeled samples from the student network. After the parameters of $F_s$ updated, $F_t$ is updated as following:
\begin{equation}
    \boldsymbol{\theta}_t^j\gets \alpha \boldsymbol{\theta}_t^j + (1-\alpha) \boldsymbol{\theta}_s^j
    \label{loss2}
\end{equation}
where $\{\boldsymbol{\theta}_t^j\}$ and $\{\boldsymbol{\theta}_s^j\}$ are the parameters of the teacher and student networks, respectively. $\alpha$ is a momentum coefficient. 


\textbf{Label Propagation.} Although weak labels can be gradually improved with the proceeding of iteratively optimizing for the teacher network, there exists some unavoidable label errors that cannot be refined and would be overfitted as the training proceeds. As the teacher and student networks have different learning ability, they are able to provide different valuable supervision information to further assist in refining weak label errors. Thus, a label propagation strategy is proposed to exploit information interaction between the two networks. Specifically, before each training generation, we employ both the teacher and student networks to obtain the hard predictions (hard weak labels) of the unlabeled set $\boldsymbol{X}^u$, denoted as $\{\hat{\boldsymbol{y}}_i^{u,t}\}, \{\hat{\boldsymbol{y}}_i^{u,s}\} \in \{0,1\}$. A label similarity matrix $\boldsymbol{C} \in \mathbb{R}^{2 \times 2}$ is then calculated via Intersection over Union (IoU) criterion between the predicting results of unlabeled sets from the two networks:
\begin{equation}
    \boldsymbol{C}_{m,n}=\frac{|\{ \boldsymbol{X}^u: \hat{ \boldsymbol{y}}_i^{u,s} = m \cap \hat{ \boldsymbol{y}}_i^{u,t} =n\}|}{|\{ \boldsymbol{X}^u: \hat{ \boldsymbol{y}}_i^{u,s} = m \cup \hat{ \boldsymbol{y}}_i^{u,t} =n\}|}
\end{equation}
where $m,n \in \{0,1\}$, and $|\cdot|$ counts the number of the unlabeled samples. $\boldsymbol{C}_{m,n}$ measures the consensus between the label class $m$ in the student network and the label class $n$ in the teacher network. 
$\boldsymbol{C}$ is then normalized such that each row sums to $1$.
After the estimated weak label similarity matrix obtained, the complementary weak supervision information from the student network can be propagated to the teacher network to refine the weak labels. The soft predictions $\{\boldsymbol{y}_i^u = \boldsymbol{p}_i^{u,t}\}_{i=1}^{N_u}$ from the teacher network are purified:
\begin{equation}
    [1-\boldsymbol{y}_i^u,\boldsymbol{y}_i^u]\gets(1-\beta)[1-\boldsymbol{y}_i^{u,s}, \boldsymbol{y}_i^{u,s}]\boldsymbol{C}+ \beta[1-\boldsymbol{y}_i^u,\boldsymbol{y}_i^u]
    \label{lp}
\end{equation}
where $\boldsymbol{y}_i^{u,s} = \boldsymbol{p}_i^{u,s} \in [0,1]$ denotes the soft prediction of the unlabeled sample from the student network. The refined weak label $\boldsymbol{y}_i^u$ is employed in Eq.~\ref{loss} and Eq.~\ref{loss2} to optimize the mean-teacher framework.

\textbf{Label Reliability Estimation.} Apart from utilizing label propagation to refine noisy weak labels, we propose to evaluate the credibility of the weak labels and suppress the contribution of samples with error-prone weak labels in the loss function for alleviating the negative effect. Based on the uncertainty theory \cite{zheng2021rectifying,kendall2017uncertainties}, a network has uncertainty on its predicted result of an input sample. The more uncertain result a network predicts, the less reliable its output is, and thus is more likely to generate incorrect weak labels. The uncertainty can be utilized to softly assess the credibility of the weak labels. We evaluate the credibility (uncertainty) of weak labels based on the output consistency (inconsistency) between the teacher and student networks. A straightforward strategy is to calculate the distance between the reinforced semantic features $\boldsymbol{f}_{c}^{u,t}$ and $\boldsymbol{f}_c^{u,s}$ from the two networks as the consistency. Nevertheless, this strategy does not consider the global distribution information in the whole unlabeled set. We thus exploit the class likelihood vectors $\boldsymbol{q}_i^{u,s}$, $\boldsymbol{q}_i^{u,t}$ to evaluate the consistency. For each unlabeled sample $\boldsymbol{x}_i^u$ and its reinforced semantic feature $\boldsymbol{f}_{c}^{u,s}$ from the student network, we calculate its similarity to 2-class center vector $\bar{\boldsymbol{f}}_c^{u,s}$ to obtain the class likelihood vector $\boldsymbol{q}_i^{u,s}$ from the student network as follows:
\begin{equation}
    \boldsymbol{q}_i^{u,s} = \text{Softmax}(\bar{\boldsymbol{f}}_c^{u,s} \cdot \boldsymbol{f}_{c,i}^{u,s})
\end{equation}
where $\bar{\boldsymbol{f}}_c^{u,s} \in \mathbb{R}^{2\times d}$ is the center vector of 2 classes (fake or real) on the unlabeled set, which is calculated by averaging all reinforced semantic features of the unlabeled samples predicted fake or real on the unlabeled set. Similarly, the class likelihood vector $\boldsymbol{q}_i^{u,t}$ from the teacher network is obtained. Afterwards, we employ the Maximum Mean Discrepancy (MMD) with Gaussian kernel \cite{gretton2012kernel} to measure the distance between the two class likelihood vectors from the two networks as the uncertainty $u_i$ of the unlabeled sample $\boldsymbol{x}_i^u$:
\begin{equation}
    u_i = \text{MMD}(\boldsymbol{q}_i^{u,t},\boldsymbol{q}_i^{u,s})
\end{equation}
The credibility of corresponding estimated weak label is calculated as $\omega_i=\exp(-u_i)$ for  enabling stable optimizing. For an unlabeled sample $x_i^u$ with high uncertainty $u_i$,  a smaller weight $\omega_i$ is employed to reduce its negative contribution in the loss function. Thus, we incorporate $\omega_i$ into the binary cross entropy loss in Eq.~\ref{loss}, and define the credibility-aware cross entropy loss as follows:
\bgroup
\setlength\abovedisplayskip{0pt}
\begin{equation}
    \begin{aligned}
    \mathcal{L}_{u} = \frac{1}{B}\sum_{i=1}^B -\omega_i [\boldsymbol{y}_i^u \log \boldsymbol{p}_i^{u,s}  
    + (1-\boldsymbol{y}_i^u) \log (1-\boldsymbol{p}_i^{u,s}) ]
    \end{aligned}
\end{equation}
\egroup
The credibility-aware cross entropy loss is finally used to optimize the framework on the unlabeled samples. 

\section{Experiments}


\begin{table*}[t]
    \centering
    \caption{Performance comparison of the state-of-the-art methods on our Wechat dataset.}
    \bgroup
    \small
    \setlength\tabcolsep{4.5pt}
    \begin{tabular}{c|c|cc|ccc|ccc}
        \hline
        \multirow{2}{*}{\textbf{Settings}}        & \multirow{2}{*}{\textbf{Method}} & \multirow{2}{*}{\textbf{Accuracy}} & \multirow{2}{*}{\textbf{AUC-ROC}} & \multicolumn{3}{c|}{\textbf{Fake News}} & \multicolumn{3}{c}{\textbf{Real News}}                                        \\
        \cline{5-10}
                                         &                         &                           &                          & \textbf{Precision}                      & \textbf{Recall}                        & \textbf{F1}     & \textbf{Precision} & \textbf{Recall} & \textbf{F1}     \\
        \hline
        \multirow{5}{*}{Supervised}      & LIWC-SVM               & 58.0                      & 60.9                     & 57.9                           & 53.6                          & 55.6   & 56.8      & 62.4   & 59.5   \\
                                         & LSTM                 & 74.1                      & 80.5                     & 87.2                           & 56.1                          & 68.3   & 67.5      & \textbf{92.1}   & 77.9   \\
                                         & CNN                   & 74.3                      & 82.8                     & 86.2                           & 58.9                          & 70.0   & 69.2      & 89.7   & 77.5   \\
                                         & EANN              & 76.1                      & 79.4                     & 86.5                           & 62.9                          & 72.8   & 71.4      & 89.3   & 79.3   \\
                                         & DualEmotion  & 77.4 & 81.7 & 84.9 & 66.4 & 74.5 & 73.8 & 88.4 & 80.4 \\
                                         & LNMT(S)                & $78.7$                    & $82.9$                   & $85.2$                         & $67.7$                        & $75.4$ & $73.2$    & $89.7$ & $80.6$ \\
        \hline
        \multirow{3}{*}{Semi-Supervised} & $\text{LSTM}+\text{Semi}$    & $75.6$                    & $83.7$                   & $84.9$                         & $61.7$                        & $71.4$ & $70.4$    & $89.6$ & $78.9$ \\
                                         & $\text{CNN}+\text{Semi}$     & $75.4$                    & $84.2$                   & $85.1$                         & $62.8$                        & $72.3$ & $71.1$    & $88.0$ & $78.7$ \\
             & LNMT(SS)   & $79.6$                    & $84.5$                   & $82.8$                         & $70.2$                        & $75.9$ & $74.8$    & $89.0$ & $81.2$ \\                                         
        \hline
        \multirow{3}{*}{Weakly-Supervised} & $\text{LSTM}+\text{LP}+\text{LR}$    & 77.5                      & 84.6                     & 86.5                           & 66.2                          & $75.1$ & 74.3      & 88.8   & 81.0   \\
                                         & $\text{CNN}+\text{LP}+\text{LR}$     & 78.1                      & 84.9                     & 86.2                           & 68.0                          & $76.0$ & 75.7      & 88.2   & 81.5   \\
                                          & $\text{WeFEND}$             & 79.9                     & 84.3                    & 85.9                        & 72.3                      & 78.5 & 75.9      & 87.3   & 81.2   \\
                                           & $\text{LNMT}$             & \textbf{82.6}                      & \textbf{87.1}                    & \textbf{88.3}                         & \textbf{74.8}                        & \textbf{81.0} & \textbf{77.5}      & {90.4}   & \textbf{83.5}   \\
        \hline
       \end{tabular}
    \egroup
    \label{table.sota}
\end{table*}

\subsection{Dataset and Evaluation Metrics}
The Wechat dataset \cite{wang2020weak} is collected from WeChat’s Official Accounts. In order to promote fake news detection, the WeChat’s Official Account encourages users to report suspicious articles, and write feedback messages to explain the reasons. Note that the publicly released version of the Wechat dataset does not match the used version in the method WeFEND \cite{wang2020weak}. WeFEND does not provide the information about the used version of the Wechat dataset. In our experiments, we thus randomly select samples from the publicly released version of the Wechat dataset to create 5 sub-datasets (Following the same setting of the used version of the Wechat dataset) for evaluation, each of which contains 2,440 news articles for training, 1,000 news articles for validation, 1,740 news articles for testing and 22,981 articles for the unlabeled set. Training, validation and test sets contain an equal amount of fake and real news articles, respectively. Detailed statistics of our dataset can be found in \textit{Supplementary Material}. We average the performance on the five sub-datasets as the final results. Following previous methods, we adopt Accuracy, AUC-ROC and class-wise Precision, Recall and $F_1$-Score as the evaluation metrics.

\subsection{Implementation Details}

We adopt Chinese Word Vectors to initialize the word embeddings in the  baseline model, which contains vectors of length 300 \cite{li-etal-2018-analogical}. We set the input and output dimensions of linear projections in all feed-forward networks and MHA layers ($\mathbb{R}^d$, $\mathbb{R}^{d_{model}}$) to 300. Following the work \cite{zhang2021mining}, several emotion sources are adopted to extract robust emotion vector ($\mathbb{R}^{d_{emo}}$) of length 47, which is composed of emotion lexicon and intensity features, sentiment score and other auxilary features. We adopt the Affective Lexicon Ontology to extract emotion lexicon and intensity features with length 29, employ the dictionary HowNet to calculate sentiment score with length 1 and extract other auxiliary features with length 17. The number of MHA layers are set as $L_w=2$ and $L_r=4$. We use a maximum sequence length of 40 and 100 for token-level and report-level MHA layers in the baseline model, respectively. In the first training stage, the input batch size of the baseline model is set to 48, which is trained for 200 epochs by using Adam optimizer. We adopt a multi-stage learning rate schedule, with the learning rate linearly increasing from 0 to 3e-5 in the first 8000 iterations, and decreasing linearly to 3e-7 in the remaining iterations. In the second training stage, we set the momentum coefficient $\alpha=0.999$ in Eq.~\ref{loss2},  label propagating rate $\beta=0.9$ in Eq.~\ref{lp}. We adopt Adam optimizer to train the mean teacher framework with a fixed learning rate 2e-6. The framework is optimized with batch size of 64, for 15 epochs.


\subsection{Comparison to State-of-the-Art Methods}

\begin{table}[!t]
    \centering
     \caption{Effectiveness of the emotion-aware encoder within the baseline model.}
     \small
    \begin{tabular}{c|cc}
        \hline
        \textbf{Model}    & \textbf{Accuracy} & \textbf{AUC-ROC} \\\hline
        w/o emotion & 77.4     & 81.7    \\
        emotion & 78.7     & 82.9    \\\hline
    \end{tabular}
   \label{table.ablation-net}
   \vspace{-0.30cm}
\end{table}

\begin{table}[!t]
    \centering
    \caption{Effectiveness of label propagation and label reliability estimation within the mean teacher framework.} 
    \small
    \bgroup
    \setlength\tabcolsep{5pt}
    \begin{tabular}{c|cc|cc}
        \hline
                          \textbf{Model} & \textbf{LP} & \textbf{LR} & \textbf{Accuracy} & \textbf{AUC-ROC} \\\hline
        Stage 1                               & -           &-         & 78.7     & 82.9    \\\hline
        \multirow{3}{*}{Stage 1 + Stage 2} &     $\times$     &  $\times$             & 79.5     & 84.0    \\
                                             & \checkmark  & $\times$         & 80.8     & 85.4    \\
                                             & \checkmark  & \checkmark  & 82.6     & 87.1    \\\hline
    \end{tabular}
    \egroup
    \label{table.ablation}
    \vspace{-0.30cm}
\end{table}


We compare LNMT with several fake news detection methods under supervised, semi-supervised and weakly-supervised settings on our Wechat dataset in Table~\ref{table.sota}, including: LIWC-SVM \cite{liwc2015},  LSTM \cite{ruchansky2017csi}, CNN \cite{EANN}, EANN \cite{EANN}, DualEmotion\cite{zhang2021mining}, WeFEND\cite{wang2020weak}. The details of these compared methods and LNMT under the three settings can be found in \textit{Supplementary Material}. It can be observed that LNMT achieves the best performance in terms of Accuracy, AUC-ROC, Precision, Recall and $F_1$ metrics. \textbf{In the supervised setting}, all the methods directly utilize the labeled samples to train the models and predict the results on the testing set. Among the supervised learning approaches, deep learning based ones such as LNMT(S), DualEmotion are superior to the traditional method LIWC-SVM, which implies that neural network is more capable in capturing representative features. Particularly, LNMT(S) employs the transformer structure to capture semantic dependencies at the token-level and report-level, and incorporates emotion information to improve the capacity. This design derives more robust features and achieves the best result in supervised setting. \textbf{In the semi-supervised setting}, all methods utilize external unlabeled data and entropy minimization loss \cite{grandvalet2005semi} for training. Semi-supervised learning greatly increases the amount of training samples, and brings a performance boost in LSTM+Semi, CNN+Semi and LNMT(SS). Furthermore, LNMT(SS) improves the second best compared method CNN+Semi by 4.2\% in Accuracy and 0.3\% AUC-ROC, respectively. \textbf{In the weakly-supervised setting}, LSTM+LP+LR, CNN+LP+LR and our LNMT not only utilize unlabeled samples as supplement, but also tackle noises in the weak labels by using label propagation and label reliability estimation strategies. Such strategies ensure more reliable training samples with refined weak labels during model optimization. The experiment results show greater improvements in performance than these models with semi-supervised learning scheme, achieving a boosting of 3\%-4\% in Accuracy and AUC-ROC. The improvement verifies the effectiveness of the proposed two strategies. Moreover, LNMT improves the second best weakly-supervised method WeFEND by 2.7\% in Accuracy and 2.8\% in AUC-ROC, which further demonstrates the superiority of LNMT.

\subsection{Ablation Study}


\textbf{Analysis of the baseline model.} In Table~\ref{table.ablation-net}, we investigate
the influence of the emotion-aware encoder in the baseline model. ``w/o emotion'' denotes the baseline model without using emotion-aware encoder, and ``emotion'' denotes the whole the baseline model. It can be observed that without using emotion-aware encoder leads to 1.3\% and 1.2\% drop in Accuracy and AUC-ROC respectively, which confirms that exploring emotional association between news article and reports can enhance the capacity of the learned features.

\textbf{Analysis of LP and LR.} We conduct the experiments to analyze the influence of label propagation and label reliability estimation in the mean teacher framework. As shown in Table~\ref{table.ablation}, compared to the baseline model, the mean teacher framework without using LP and LR achieves 79.5\% and 84.0\% in Accuracy and AUC-ROC, obtaining 0.8\% and 1.1\% improvement in Accuracy and AUC-ROC, respectively. This improvement indicates the teacher and student networks have different learning ability, which help initially correct weak label errors. By applying LP to the mean teacher framework, the model performance is improved by 1.3\% Accuracy and 1.4\% AUC-ROC, which verifies the effectiveness of the label propagation strategy to exploit information interaction between the two networks for depressing noises in weak labels. By further applying LR to the mean teacher framework, the model performance is improved by 1.8\% Accuracy and 1.7\% AUC-ROC, indicating the effectiveness of the label reliability estimation to suppress the contribution of samples with error-prone weak labels during model training.





\section{Conclusion}
In this paper, we present a novel label noise-resistant mean teaching approach (LNMT) for weakly supervised fake news detection. LNMT automatically assigns initial weak labels to unlabeled samples based on the semantic correlation and  emotional associations between news contents and the reports. Moreover, LNMT establishes a mean teacher framework with different learning ability and information interaction. The label propagation and label reliability estimation within the framework is able to refine the weak labels and alleviate the negative effect of noisy weakly labeled samples within the loss function. Extensive experiments show that the proposed LNMT achieves significant improvement over the state-of-the-art methods.

\bibliographystyle{named}
\bibliography{ijcai22}

\end{document}